\documentclass[conference]{IEEEtran}
\IEEEoverridecommandlockouts
\usepackage{cite}
\usepackage{amsmath,amssymb,amsfonts}
\usepackage{algorithmic}
\usepackage{graphicx}
\usepackage{textcomp}
\usepackage{xcolor}
\usepackage{algorithm}
\usepackage{algorithmic}
\usepackage{booktabs}
\usepackage{newfloat}
\usepackage{listings}
\usepackage{times}
\usepackage{latexsym}
\usepackage{xcolor}
\usepackage{lipsum}
\usepackage{amsmath}
\usepackage{amssymb}
\usepackage{defs}
\usepackage{multirow}
\def\BibTeX{{\rm B\kern-.05em{\sc i\kern-.025em b}\kern-.08em
    T\kern-.1667em\lower.7ex\hbox{E}\kern-.125emX}}
\begin{document}

\title{
Navigating the Structured What-If Spaces: Counterfactual Generation via Structured Diffusion
}

\author{\IEEEauthorblockN{Nishtha Madaan\textsuperscript{*}}
\IEEEauthorblockA{\textit{Indian Institute of Delhi, India} \\
nishtha.madaan@iitd.ac.in}
\thanks{*Nishtha Madaan is a researcher at IBM Research and this work was done as a part of PhD research at IIT Delhi.}
\and
\IEEEauthorblockN{Srikanta Bedathur}
\IEEEauthorblockA{\textit{Indian Institute of Delhi, India} \\
srikanta@cse.iitd.ac.in
}}

\maketitle

\begin{abstract}
Generating counterfactual explanations is one of the most effective approaches for uncovering the inner workings of black-box neural network models and building user trust. While remarkable strides have been made in generative modeling using diffusion models in domains like vision, their utility in generating counterfactual explanations in structured modalities remains unexplored. In this paper, we introduce \textit{Structured Counterfactual Diffuser} or SCD, the first plug-and-play framework leveraging diffusion for generating counterfactual explanations in structured data. SCD learns the underlying data distribution via a diffusion model which is then guided at test time to generate counterfactuals for any arbitrary black-box model, input, and desired prediction. Our experiments show that our counterfactuals not only exhibit high plausibility compared to the existing state-of-the-art but also show significantly better proximity and diversity.
\end{abstract}

\section{Introduction}

% Grand Challenge
As AI models become more capable and widespread, the issue of trust becomes critical \cite{DoshiVelez2017TowardsAR}. While traditional software is transparent---allowing tracing its control flow and easily resolving trust concerns---modern AI is built upon neural networks that are not transparent. Their underlying control flow is not understood, making it difficult to trust in high-risk settings such as loan or hiring decisions. Although the remarkable power and flexibility of neural networks have allowed building systems that achieve capabilities not possible with traditional software alone \cite{OpenAI2023GPT4TR, Ramesh2022HierarchicalTI}, this lack of transparency and trust becomes a significant hurdle in realizing the full potential of neural networks \cite{ribeiro2016model, lundberg2017unified, wachter2017counterfactual}.

% Zero-in on the exact line of research -- all about Wachter and no more.
To address concerns about trust, one needs to answer \textit{why} a model behaves in a certain way. One of the most promising directions to answer this is via \textit{what-if} scenarios or counterfactuals \cite{wachter2017counterfactual}.  For instance, consider a model which declines a loan for \texttt{[Female, Earns \$100K]}. To answer \textit{why}, it is of interest to discover counterfactuals for which the same model approves loans.
For instance, if the model approves the loan for a counterfactual instance \texttt{[Male, Earns \$100K]}, this suggests that the model may be making decisions based on potentially problematic criteria, prompting model developers to investigate and fix the problem. Additionally, counterfactuals can also provide actionable insights to the end-users on how to achieve a different outcome \cite{mothilal2020explaining}. In our previous example, if the model approves the loan for a counterfactual instance \texttt{[Female, Earns \$110K]}, it explains what the applicant might need to do to obtain approval.

% PARA 3: ONE LINE -- HIGHLIGHT PLAUSIBILITY PROBLEM + RW
While \cite{wachter2017counterfactual} originally introduced the idea of counterfactual explanations, the idea has gained significant attention in recent years \cite{mothilal2020explaining, karimi2019mace, yang2022mace, ross2020explaining, madaan2021generate}. Ideally, counterfactuals should possess the following characteristics: \textit{1)} they should maintain \textit{proximity} to the original input, \textit{2)} they should attain the desired counterfactual label to ensure its \textit{validity}, \textit{3)} they should be \textit{diverse} and capture a wide range of distinct scenarios and \textit{4)} they should be \textit{plausible}. While proximity, validity, and diversity criteria have been studied extensively, there has been little focus on the plausibility of the generated counterfactuals, i.e., ensuring that the generated counterfactuals are realistic and conform to the underlying data distribution. Previous works have approached plausibility in a minimal sense, e.g., enforcing values to lie in legal ranges or applying user-designed constraints \cite{mothilal2020explaining, karimi2019mace}. 
% This raises the question \textit{``Would learning a rich data-driven model of the underlying data distribution help generate more plausible counterfactuals?"}

Recently, in the visual domain, diffusion models \cite{ho2020denoising} have been successfully used to acquire the underlying data distribution for generating plausible counterfactual explanations \cite{augustin2022diffusion, Jeanneret2022DiffusionMF, Sanchez2022DiffusionCM, 2023DiffusionbasedVC}. However, in the domain of tabular or structured data, counterfactual explanation methods have largely ignored these recent advances in diffusion modeling raising another important question: \textit{``Can diffusion models, which are known for their remarkable generation capabilities in vision, help generate high-quality plausible counterfactuals in the structured domain?"}

% \textcolor{red}{[Emphasize that it is not just CV.]}

% However, most counterfactual explanation methods that do seek to acquire such underlying data distribution for generating plausible counterfactuals have overlooked the domain of structured data and largely remained confined to the visual \cite{goyal2019counterfactual, Lang2021ExplainingIS, Mertes2020GANterfactualCounterfactualEF} and language domain \cite{ross2020explaining, madaan2021generate, madaan2023counterfactual}. Recently, in the visual domain, diffusion models \citep{ho2020denoising} have been successfully used to generate counterfactual explanations \cite{augustin2022diffusion, Jeanneret2022DiffusionMF, Sanchez2022DiffusionCM, 2023DiffusionbasedVC} raising another important question: \textit{``Can diffusion models, which are known for their remarkable generation capabilities, help generate high-quality counterfactuals for the structured domain?"}

% These systems lack a rich data-driven model of the underlying data distribution to constraint the counterfactuals to be plausible. 
% - Existing approaches take a very slap-dash approach towards plausibility ---
% - What we need is some kind of TABLE-MODEL 
% % PARA 4: WE PROPOSE ... IT WORKS LIKE THIS
% In this work, SCD
% - DIFFUSION
% - ITS GOOD CANDIDATE BECAUSE IMAGE
% - Another way to look at our work is that its first application of diffusion --- 
% PARA 5: IN EXPERIMENTS ....

To answer this question, in this work, we propose a novel counterfactual explainer called \textit{Structured Counterfactual Diffuser} or \textit{SCD}. SCD is the first plug-and-play framework leveraging diffusion modeling for generating counterfactual explanations for structured data. SCD works by learning the underlying data distribution via a diffusion model \cite{li2022diffusion, ho2020denoising}. At test time, the diffusion model is used to perform guided iterative denoising to generate counterfactuals for any given input and black-box model in a plug-and-play manner. In experiments, we show that our counterfactual explainer not only exhibits high plausibility compared to the state-of-the-art approaches but also shows significantly better proximity and diversity scores of the generated counterfactuals. In our analysis, we also find that our method, due to its unique stochastic denoising process, does not require explicit incentives to generate diverse counterfactuals, unlike the previous counterfactual explainers for structured data.

\begin{figure*}[t]
    \centering
    \includegraphics[width=\textwidth]{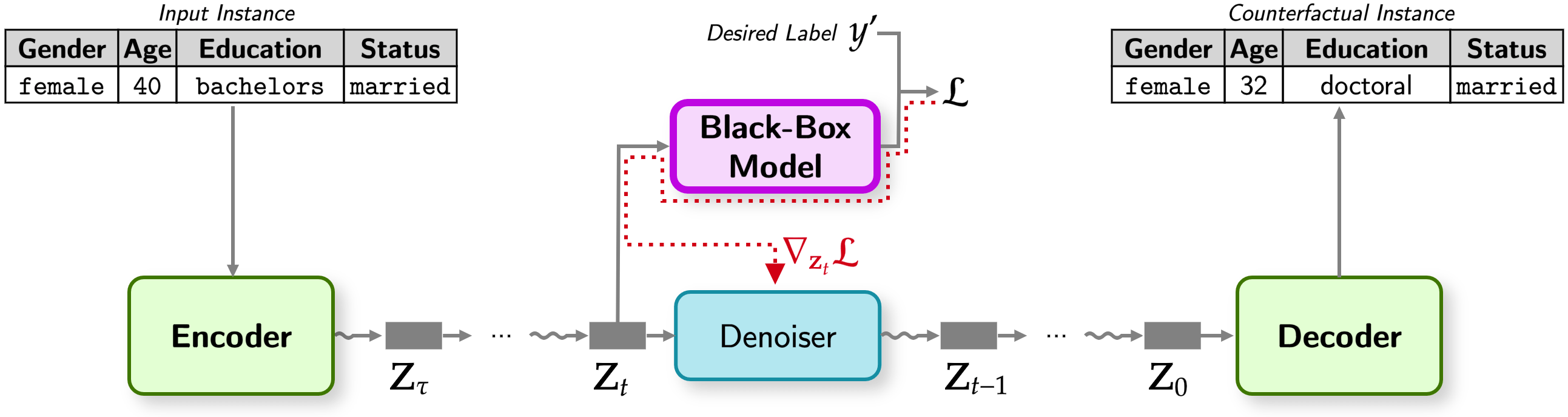}
    \caption{\textbf{Overview of Our Counterfactual Generation Process.} The process starts by encoding the given human-readable instance or row into an embedding by performing a look-up on a dictionary of learned embeddings. Next, we iteratively apply denoising steps while incorporating the gradient information from the given black-box model to minimize the disparity between the model's prediction and the desired label. At the end of the denoising process, we obtain an embedding which is then decoded via a reverse look-up on the dictionary to obtain the counterfactual instance.}
    \label{fig:one}
\end{figure*}

% ...

\section{Preliminaries}

\textbf{Structured Data.} A table or structured data consists of rows or instances. Each instance is a tuple with a value for each column or attribute. The entire space of such instances can be described as $\cX = \cX_1 \times \ldots \times \cX_C$. Here, $C$ denotes the number of columns or attributes in the table, and each $\cX_c$ denotes the space of possible values for column $c$. For example, a possible instance from a 4-column table is $\texttt{[female, 40, doctoral, married]}$. Here, $\cX_1$ can represent gender categories, $\cX_2$ can represent the possible age values, and so forth. We will use $\bx$ to denote an instance and $\bx^c$ to denote $c$-th column or attribute within the instance.\\

\noindent
\textbf{Black-Box Model.} A black-box model is a model $f:\cX\rightarrow\cY$ that maps an input instance $\bx \in \cX$ to a label $y\in\cY$. However, the model is \textit{black-box} in the sense that its inner workings are not understood and explainability tools are required to shed light on it. In the rest of the paper, we will use the term \textit{model} and \textit{black-box model} interchangeably.

% Each instance may also be associated to a class label taking values from a space $\cY$. For the example in Fig.~\ref{fig:one}, one possible space of labels $\cY$ could be credit risk.

\subsection{Structured Counterfactual Explanations}

As highlighted by \cite{wachter2017counterfactual}, counterfactuals help identify alternative scenarios where a slight change in the original input $\bx$ to a counterfactual input $\bx'$ would have changed the outcome from $y$ to $y'$ by a black-box model $f$. By analyzing the change in prediction on counterfactual inputs, one can uncover if the model is making decisions based on potentially problematic or undesired criteria.\\

\noindent
\textbf{Counterfactual Explainer.} Formally, a counterfactual explainer can be described as a system or framework that, given an input $\bx$, a model $f$, and a counterfactual label $y'$ (where $y'$ is different from the original label $y$), produces a set of $B$ counterfactuals $\bX'$.
\begin{align*}
\bX' = \{\bx'_1, \ldots, \bx'_B\} = \text{CounterfactualExplainer}(f, \bx, y').
\end{align*}
Here, each counterfactual $\bx'_b\in \bX'$ should achieve the counterfactual label $y'$ on the given black-box model $f$ with minimal change to the original input $\bx$.\\

\noindent
\textbf{Desired Characteristics of Counterfactuals.}
There are 4 fundamental characteristics that counterfactuals in $\bX'$ should possess:
\begin{enumerate}
    \item \textit{Validity:} Should achieve the label $y'$.
    \item \textit{Proximity:} Should be close to the original input $\bx$.
    \item \textit{Diversity:} Should be diverse and not collapse to a single instance.
    \item \textit{Plausibility:} Should be plausible, i.e., should capture realistic instances from the input space.
\end{enumerate}
While \cite{wachter2017counterfactual} originally introduced the validity and proximity desiderata, \cite{mothilal2020explaining} introduced the desiderata of diversity. Plausibility, on the other hand, has not been given much attention in the community. Some existing works primarily focus on only keeping generated values within legal ranges, disregarding the complex relationships that values of various columns have \cite{karimi2019mace} or require costly user-defined plausibility constraints \cite{mothilal2020explaining}. In this work, we take a significant step forward in alleviating this concern.

\section{SCD: Structured Counterfactual Diffuser}
In this section, we present our proposed model \textit{Structured Counterfactual Diffuser} or \textit{SCD}. SCD learns a diffusion model through training on a structured dataset or table $\cD$. Via training on $\cD$, SCD learns about the underlying data distribution which enables it to generate plausible counterfactuals. Once the diffusion model is trained, SCD can be used in a plug-and-play manner to obtain counterfactual explanations for any given black-box model. We now describe SCD in detail.\\

\noindent
\textbf{Row Embedding.} To train the diffusion model, we first map the raw human-readable instances or rows $\bx$ of the table $\cD$ into embeddings. The diffusion model shall be trained to model the distribution in this embedding space. 
% Later, via reverse lookup using this learned dictionary, we shall convert these embeddings back to a human-readable format. 
We maintain a learned dictionary of embeddings $\text{Embedding}_c: \cX_c \rightarrow \mathbb{R}^d$ for each column $c$. To encode a row, we lookup the embedding for each of the $C$ columns and concatenate these embeddings to obtain a row embedding $\bz$ as follows:
\begin{align*}
    \bz = [\text{Embedding}_1(\bx^1), \ldots, \text{Embedding}_C(\bx^C)] \in \mathbb{R}^{C\times d}
\end{align*}
where $d$ is the size of the embedding per column.

\subsection{Diffusion Modeling}
Via diffusion modeling, we seek to learn a distribution $p_\ta(\bz)$ over the row embeddings. In diffusion modeling, the distribution $p_\ta(\bz)$ consists of $T$ denoising steps:
\begin{align*}
    p_\ta(\bz_0) = \int p(\bz_T) \prod_{t=T,\ldots,1} p_\ta(\bz_{t-1}|\bz_{t}, t) d \bz_{1:T}
\end{align*}
Here, $p(\bz_T)$ represents standard Gaussian, the sequence $\bz_T, \ldots, \bz_1$ consists of iteratively cleaner samples, finally producing the desired sample $\bz_0$; and $p_\ta(\bz_{t-1} | \bz_{t}, t)$ is a one-step denoising distribution. The $p_\ta(\bz_{t-1} | \bz_{t}, t)$ is parametrized in the following manner:
\begin{align*}
     \cN\left( \gamma_{1,t}\hat{\bz}_0 + \gamma_{2, t}\bz_t, \beta_t\mathbf{I}  \right)
\end{align*}
where $\hat{\bz}_0 = g_\ta(\bz_t, t)$, and the coefficients $\gamma_{1,t}$ and $\gamma_{2,t}$ are given by:
\begin{align*}
    \gamma_{1,t} = \frac{\beta_t \sqrt{\bar{\alpha}_{t-1}}}{1 - \bar{\alpha}_t}, && 
    \gamma_{2,t} = \frac{(1 - \bar{\alpha}_{t-1})\sqrt{\alpha_t}}{1 - \bar{\alpha}_t}
\end{align*}
Employing standard notations, we utilize a variance schedule $\beta_1, \ldots, \beta_T$, where $\alpha_t = 1 - \beta_t$, and $\bar{\alpha}_t = \prod_{i=1}^t (1 - \beta_i)$. We use a cosine schedule in our implementation.\\

\noindent
\textbf{Unconditional Sampling.} To obtain unconditional samples from the learned diffusion model, we start with random Gaussian noise, $\bz_T \sim p(\bz_T)$. Next, using the trained one-step denoising distribution $p_\ta(\bz_{t-1} | \bz_{t}, t)$, we iteratively denoise the samples until $\bz_0$, the desired sample, is obtained.\\

\noindent
\textbf{Learning:} The training procedure involves first introducing noise to the input $\bz_0$, creating its noisy version $\bz_t$. 
\begin{align*}
    \bz_t = \sqrt{\bar{\alpha}_t} \bz_0 + \sqrt{1 - \bar{\alpha}_t} \boldsymbol{\epsilon}_t, && \text{ where } \boldsymbol{\epsilon}_t \sim \cN(\mathbf{0}, \mathbf{I}).
\end{align*}
Subsequently, a neural network predictor is trained that takes $\bz_t$ as input and aims to predict the original input $\bz_0$ by generating a prediction $\hat{\bz}_0 = g_\ta(\bz_t, t)$.
The learning objective is $\cL_\text{diffusion}(\ta) = \cE(\hat{\bz}_0, \bz_0)$ where $\cE$ is an error function.
% , commonly chosen to be the mean-squared error (MSE) in the vision domain. However, in our tabular domain, we found it more stable to apply a cross-entropy loss instead.

\subsection{Generating Counterfactuals via Guided Diffusion}

Given the trained denoising distribution $p_\ta(\bz_{t-1} | \bz_{t}, t)$, we are now ready to generate counterfactuals for a black-box model $f$ given an input instance $\bx$ and a desired label $y'$. The process works by performing guided diffusion starting from the embedding of the given input instance. For this, we first encode $\bx$ to its row embedding $\bz \in \mathbb{R}^{C\times d}$. Since we seek to sample $B$ counterfactuals, we copy the row embedding $B$ times and stack the copies together to construct an embedding $\bZ \in \mathbb{R}^{B \times C \times d}$. Next, we add Gaussian noise to $\bZ$ to facilitate diversity among the $B$ generated samples.
\begin{align*}
    \bZ'_\tau \leftarrow \sqrt{\bar{\alpha}_\tau} \bZ + \sqrt{1 - \bar{\alpha}_t} \boldsymbol{\epsilon}_t, && \text{ where } \boldsymbol{\epsilon}_t \sim \cN(\mathbf{0}, \mathbf{I}).
\end{align*}
Next, we perform $\tau$ guided diffusion steps. We iteratively and alternatingly apply the following two steps:
\begin{enumerate}
    \item \textit{Denoising Step:} This step involves sampling $\bZ'_{t-1} \sim p_\ta(\bZ'_{t-1} | \bZ'_{t}, t)$.
    \item \textit{Guiding Step:} This step involves performing a gradient step on $\bZ'_{t-1}$ with respect to a guiding loss $\cL$ as: 
    \begin{align*}
        \bZ'_{t-1} \leftarrow \bZ'_{t-1} - \eta \nabla_{\bZ'_{t-1}} \cL
    \end{align*}
    where $\eta$ is the step size for the update. One of the things that $\cL$ measures is how well the black-box model $f$ produces the counterfactual label $y'$ on the samples $\bZ'_{t-1}$ of the current step. We describe the exact formulation of $\cL$ in detail in a later section.
\end{enumerate}
From this iterative process, we obtain a series of progressively cleaned embeddings $\bZ'_\tau, \ldots, \bZ'_0$. Next, we take the generated $\bZ'_0$, perform reverse look-up using the learned embeddings and obtain the human-readable counterfactual instances $\bX' = \{\bx'_1, \ldots, \bx'_B \}$. In Fig.~\ref{fig:one}, we illustrate this process.

\textbf{Guiding Loss:}
We now describe the terms in our guiding loss $\cL$. Following \cite{mothilal2020explaining}, we include 3 terms in our loss capturing validity, proximity, and diversity of the samples. Formally this loss can be described as:
\begin{align*}
    \cL(\bZ', \bx, f, y') &= \lambda_\text{validity}\cL_\text{validity}(\bZ', f, y')\\ &+ \lambda_\text{proximity}\cL_\text{proximity}(\bZ, \bZ')\\ &+ \lambda_\text{diversity}\cL_\text{diversity}(\bZ')
\end{align*}
\begin{enumerate}
    \item \textit{Validity Loss.} We use the cross-entropy loss of the black-box model $f$ with respect to the desired prediction $y'$ as our validity loss.
    \begin{align*}
        \cL_\text{validity}(\bZ', f, y') = \text{CrossEntropy}(f(\bZ'), \texttt{target}=y').
    \end{align*}
    \item \textit{Proximity Loss.} We use a simple L2 loss between $\bZ$ the embedding of the original input and $\bZ'$ the generated embedding at the current step of the guided diffusion.
    \begin{align*}
        \cL_\text{proximity}(\bZ, \bZ') = ||\bZ - \bZ'||^2.
    \end{align*}
    \item \textit{Diversity Loss.} We use the negative of L2 loss between all pairs of counterfactual instances 
    \begin{align*}
        \cL_\text{diversity}(\bZ') = \frac{-2}{B(B-1)} \sum_{i=1}^{B-1} \sum_{j=i+1}^B ||\bz'_i - \bz'_j||^2.
    \end{align*}
\end{enumerate}
% In experiments, we shall see that our method works effectively using only the validity loss term without requiring explicit proximity and diversity terms.
\begin{table*}[t]
% \scriptsize
\centering
  \caption{Comparison of plausibility, proximity, diversity, and validity scores of Wachter, SCD and DiCE on various datasets. For validity, proximity, and diversity scores, higher is better. For the plausibility score, lower is better since it captures the negative log-likelihood of the generated samples.}
    \label{tab:quantitative-results-scd}
\begin{tabular}{lccccccc}
\toprule
\multicolumn{1}{l}{\textbf{Dataset}} & \multicolumn{3}{c}{\textbf{Plausibility} $(\downarrow)$} & \multicolumn{3}{c}{\textbf{Proximity} $(\uparrow)$} \\
\cmidrule(lr){2-4} \cmidrule(lr){5-7}
 & \textbf{Wachter \textit{et al.}} & \textbf{DiCE} & \textbf{SCD} & \textbf{Wachter \textit{et al.}} & \textbf{DiCE} & \textbf{SCD} \\
\midrule
\textbf{Adult Income} & 108.7 & 121.0 & \textbf{21.21} & 0.685 & 0.5764 & \textbf{0.6173} \\
\textbf{UCI Bank} & 168.3 & 166.7 & \textbf{42.37} & 0.226 & 0.2141 & \textbf{0.3000} \\
\textbf{Housing Price} & 102.8 & 109.5 & \textbf{42.91} & 0.375 & 0.3055 & \textbf{0.3417} \\
\midrule
\multicolumn{1}{l}{\textbf{Dataset}} & \multicolumn{3}{c}{\textbf{Diversity} $(\uparrow)$} & \multicolumn{3}{c}{\textbf{Validity} $(\uparrow)$} \\
\cmidrule(lr){2-4} \cmidrule(lr){5-7}
 & \textbf{Wachter \textit{et al.}} & \textbf{DiCE} & \textbf{SCD} & \textbf{Wachter \textit{et al.}} & \textbf{DiCE} & \textbf{SCD} \\
\midrule
\textbf{Adult Income} & 0.002 & 0.3837 & \textbf{0.4008} & 0.9400 & \textbf{0.9776} & 0.7511 \\
\textbf{UCI Bank} & 0.041 & 0.4165 & \textbf{0.5498} & 0.9900 & \textbf{0.9686} & 0.8600 \\
\textbf{Housing Price} & 0.03 & 0.4289 & \textbf{0.5986} & 0.9999 & \textbf{0.9908} & 0.8526 \\
\bottomrule
\end{tabular}
\end{table*}
\subsection{Discussion}

Our method has multiple benefits. First, our method operates in a plug-and-play manner once the diffusion model is trained. That is, no training is required during sampling of counterfactual explanations:\begin{align*}\bX' = \text{CounterfactualExplainer}_\ta(f, \bx, y').\end{align*}
Second, our experiments shall show that our method can produce diverse samples without requiring an explicit diversity term in the guiding loss, distinguishing it from previous methods like DiCE \cite{mothilal2020explaining} which require an explicit diversity term in the loss. 
Third, our experiments shall show that our method can inherently preserve the contents of the original input without requiring an explicit proximity term in the guiding loss---another attractive aspect of our method. Fourth, not requiring proximity and diversity terms removes the burden of tuning the coefficients $\lambda_\text{proximity}$ and $\lambda_\text{diversity}$ which can be quite brittle in the previous methods.
\begin{table*}[t]
    \scriptsize
    \centering
    \caption{\textbf{Counterfactual Samples in Adult Income Dataset.} Given the input row with the original label ``$\leq50K$", we ask our method SCD and the baseline DiCE to generate counterfactual instances that flip the label to ``$>50K$" with respect to a black-box income predictor. We note that SCD generates plausible samples while DiCE struggles. Specifically, we note that DiCE creates counterfactuals containing \textit{Divorced} and \textit{Husband} within the same row which is contradictory and impossible (highlighted in \textcolor{red}{red}). In comparison, SCD creates plausible counterfactuals where the {Marital Status}, {Relationship} and Gender columns correctly conform with each other (highlighted in \textcolor{teal}{green}).}
    \label{tab:qual}
    \begin{tabular}{lllllllllllll}
        \toprule
        \textbf{Method} & \textbf{Age} & \textbf{Workclass} & \textbf{Education} & \textbf{Ed. No.} & \textbf{Marital Status} & \textbf{Occupation} & \textbf{Relationship} & \textbf{Race} & \textbf{Gender} & \textbf{Hr/W} & \textbf{Country} \\
        \midrule
        \textbf{Input} & 39 & State-gov & Bachelors & 13 & Never-married & Adm-clerical & Not-in-family & White & Male & 40 & US \\
        \midrule
        \textbf{Ours} & 31 & Self-emp-inc & Bachelors & 13 & \textcolor{teal}{Married-civ-spouse} & Adm-clerical & \textcolor{teal}{Wife} & White & Female & 40 & US \\
        & 34 & Self-emp-inc & Bachelors & 13 & \textcolor{teal}{Married-civ-spouse} & Exec-managerial & \textcolor{teal}{Husband} & White & Male & 55 & US \\
        & 39 & Federal-gov & Bachelors & 13 & \textcolor{teal}{Married-civ-spouse} & Prof-specialty & \textcolor{teal}{Husband} & White & Male & 40 & US \\
        \midrule
        \textbf{DiCE} & 39 & State-gov & Bachelors & 16 & \textcolor{red}{Divorced} & Transport-moving & \textcolor{red}{Husband} & White & Male & 40 & US \\
        & 39 & State-gov & Bachelors & 16 & \textcolor{red}{Divorced} & Transport-moving & \textcolor{red}{Husband} & AME & Male & 40 & US \\
        & 39 & Without-pay & Some-college & 16 & \textcolor{red}{Divorced} & Transport-moving & \textcolor{red}{Husband} & White & Male & 40 & US \\
        \bottomrule
    \end{tabular}
\end{table*}

% Third, As our experiments will demonstrate, our approach doesn't heavily rely on such a term. Furthermore, our model instinctively preserves the content of the input sample because the diffusion process commences from the given input. Consequently, it's improbable, based on our experiments, that the generated samples will diverge significantly from the provided input. This inherent characteristic eliminates the need to meticulously tune losses for diversity and proximity, a crucial yet delicate aspect of DiCE. Reinforcement Learning can be effortlessly extended to non-differentiable classifiers by employing a REINFORCE formulation of the guiding loss 
% $\cL$.

% Our method is plug and play because ...

% Our model is naturally a diverse sampler ... unlike DiCE which explicitly needs a diversity term in the loss, ours, we we shall see in our experiments, its not so important to have diversity term.

% Our model naturally preserves content with the input sample, because we start the diffusion anyway from the given input ... so it is unlikely, as we shall see in our expermients, that the generated samples will be too far from the given input ...

% As such, it frees us from tuning the losses to create diversity and proximity --- which was an essential and sensitive aspect of DiCE.

% \textbf{Reinforcement Learning} .... Naturally extended to non-differentiable classifiers by adopting a REINFORCE formulation of the guiding loss $\cL$ ... 

\section{Related Work}

\textbf{Explainable AI.} Explainable AI (XAI) has received significant attention over the past few years \cite{molnar2020interpretable}. Several methods seek saliency maps as a way of explanation \cite{simonyan2013deep, han2020explaining}. The notion of generating explanations has been well studied in image domain \cite{hendricks2018grounding}, \cite{goyal2019counterfactual}, and
\cite{van2021interpretable}.  Several methods focus on perturbing the input instance, however, these perturbations are not optimized to achieve a counterfactual prediction under the given black-box model \cite{li2016understanding, feng2018pathologies, checklist:acl20}. Another line of work on generating adversarial instances of an input instance has been well studied \cite{ebrahimi2017hotflip, ribeiro2018semantically, iyyer-etal-2018-adversarial, jia2017adversarial}. However, unlike ours, these methods are not concerned with the plausibility of the samples. Other classes of methods include approximation-based ones which learn local or global decision boundaries to generate explanations \cite{ribeiro2016model, lundberg2017unified, ribeiro2018anchors}. However, these are not counterfactual explainers. Another way to achieve interpretability has been to introduce disentanglement within the neural network layers \cite{Higgins2016betaVAELB}. However, this approach does not seek to explain existing black-box models. \\

% \textcolor{red}{[CITE:RUSSEL, CITE:FACE]}
% \cite{face}
% \cite{russell}

\noindent
\textbf{Counterfactual Explanations.} For the tabular domain, various studies have pursued counterfactual explanations \cite{wachter2017counterfactual, mothilal2020explaining, yang2022mace, karimi2020model, Guidotti2019FactualAC}. However, none of them directly and properly tackle the problem of generating plausible counterfactuals. \cite{face} propose a technique to select counterfactual samples from the training set and show the applicability on synthetic datasets.  However, their focus is not on generating completely new counterfactuals that don't occur within the training set. 
In the image domain, several works attempt to generate counterfactuals using diffusion models \cite{augustin2022diffusion, Jeanneret2022DiffusionMF, Sanchez2022DiffusionCM, 2023DiffusionbasedVC}. This is another line of works focusing on contrastive explanations \cite{Dhurandhar2019ModelAC, Jacovi2021ContrastiveEF}, however, these do not leverage diffusion modeling, like ours.
However, while these are based within the image domain, the utility of diffusions models for counterfactual explanation in the tabular domain has remained unexplored.
In the language domain, there has been a significant number of works for counterfactual generation \cite{wu2021polyjuice, madaan2021generate, madaan2023counterfactual, ross2020explaining, Boreiko2022SparseVC, Howard2022NeuroCounterfactualsBM}. However, these have primarily relied on auto-regressive LLMs and not diffusion models. Although \cite{li2022diffusion} pursues diffusion-based language modeling, it does not pursue the task of counterfactual explanation and also does not deal with the tabular domain.
Additionally, there has also been interest in the domain of search and retrieval for generating counterfactual explanations \cite{Xu2023CounterfactualEF}.

\section{Experiments}
\label{expts}

\textbf{Datasets.} In experiments, we evaluate the quality of generated counterfactuals on three datasets:
\begin{enumerate}
    \item \textbf{Adult Income Dataset} \cite{frank2010uci}. This dataset contains educational, demographic, and occupancy information of individuals. We use the following features: hours per week, education level, occupation, work class, race, age, marital status, and sex. These are selected following the pre-processing approach of \cite{zhu2016predicting}. 
    \item \textbf{UCI Bank Dataset} \cite{miscbankmarketing222}. This dataset contains the marketing campaigns of a banking institution. 
    \item \textbf{Housing Price Dataset} \cite{pace1997sparse}. This dataset contains information regarding the demography (income, population, house occupancy) in the districts of California, the location of the districts (latitude, longitude), and general information regarding the house in the districts (number of rooms, number of bedrooms, age of the house). \\
\end{enumerate}
\begin{figure*}[t]
    \centering
    \includegraphics[width=0.9\textwidth]{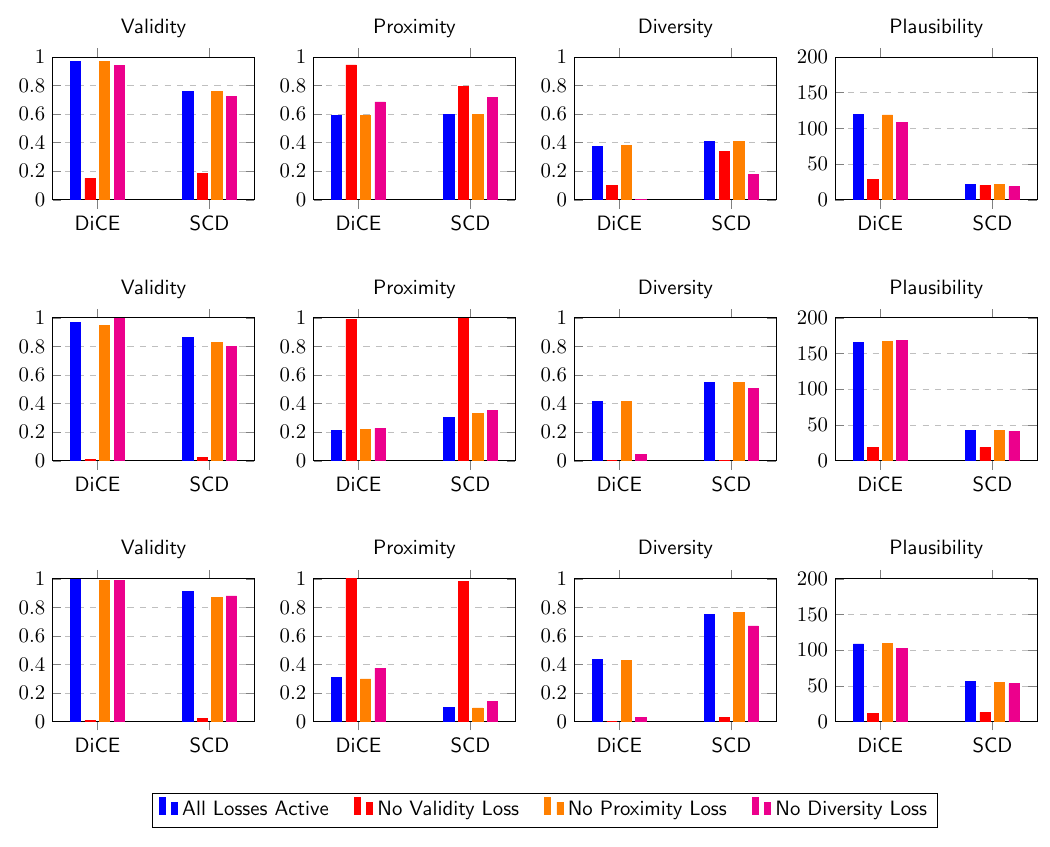}
    \caption{\textbf{Ablation of Losses.} We perform a comparison of models on ablations of the guiding loss: \textit{1)} all losses are active, \textit{2)} no validity loss, \textit{3)} no proximity loss, and \textit{4)} no diversity loss. This comparison is done on Adult Income Dataset (\emph{top}), UCI Bank Dataset (\emph{middle}) and Housing Price Dataset (\emph{bottom}). We note that when diversity loss is dropped, the performance of the baseline DiCE suffers while our model SCD maintains good diversity, proximity, and validity. In general, our model SCD maintains a very high plausibility in all scenarios relative to DiCE.}
    \label{fig:ablation}
\end{figure*}
\noindent
\textbf{Black-Box Model.} For each dataset, we train a classifier to act as the black-box model that a counterfactual explainer would seek to explain. The architecture is a simple 2-layer MLP that takes the concatenated embeddings of columns of a row as input and tries to predict a class label. For each dataset, the classification task that the black-box model is trained to perform is as follows: \textit{1)} \textit{Adult Income Dataset:} Given a row as input, the black-box model predicts whether the income exceeds $50$K per year or not. \textit{2)} \textit{UCI Bank Dataset:} Given a row describing attributes of a client, the black-box model predicts if the client will subscribe to a term deposit or not. \textit{3)} \textit{Housing Price Dataset:} Given a row as input, the black-box model predicts whether the house price is greater than \$200K or not. \\

\noindent
\textbf{Baselines.} 
% \textcolor{red}{[Emphasize that there are not many baselines.]}
We compare our model with two baseline counterfactual explainers for structured datasets: \textit{1)} DiCE, the current state-of-the-art, and \textit{2)} Wachter \textit{et al.} \cite{wachter2017counterfactual}.
These work by encoding the given row to a vector of per-column one-hot embeddings. To generate counterfactuals, they apply Stochastic Gradient Descent (SGD) to minimize a loss having terms focusing on validity, proximity, and diversity (in DiCE).  These baselines provide the most comprehensive evaluation of the proposed model since, like ours, it also leverages gradient-based dynamics to generate counterfactuals. Although it might appear that the number of compared baselines is small, we highlight that this line of research, although important, is still in its infancy and the two baselines we compare with are the most relevant with respect to our contribution.
% Another baseline is \cite{wachter2017counterfactual}. It works by generating a perturbation to the input text by focusing on validity and proximity. 
% It then applies Stochastic Gradient Descent (SGD) on $\bz$ to obtain a counterfactual embedding $\bz'$ by minimizing a loss $\cL(\bz)$. Finally, the $\bz'$ is converted back to a tabular instance by inverting the encoding process. In the loss $\cL$, DiCE concentrates on three fundamental aspects, as expressed by the loss function $\cL$:
% \begin{enumerate}
%     \item \textit{Validity:} Ensuring that the counterfactual $\bz'$ possesses validity by adhering to the desired label $y'$.
%     \item \textit{Proximity:} Ensuring proximity to the original input $\bx$.
%     \item \textit{Diversity:} Ensuring diversity within the counterfactuals generated.\\
% \end{enumerate}
% However, it is important to note that DiCE, while pursuing in these dimensions, falls short in one crucial aspect: \textit{plausibility}. That is, generated counterfactuals by DiCE lack a constraint ensuring their plausibility, i.e., they rarely represent feasible or realistic instances within the input space.

\subsection{Metrics}
% \textcolor{red}{\lipsum[1]}
We consider the following metrics for evaluating the generated counterfactuals. 

\begin{enumerate}
\item \textbf{Validity Score.}
We compute the validity score of the generated counterfactuals in $\bX'$ by checking if they result in the desired label with respect to the black-box model. 
\begin{align*}
    \text{Validity Score} = \frac{1}{B}\sum_{b=1}^B \mathbb{I}(y' == f(\bx'_b))
\end{align*}
where $\mathbb{I}(\cdot)$ is an indicator function that takes a value 1 if its input is true else 0.
\item \textbf{Proximity Score.}
We compute proximity score as the mean of distances between the generated counterfactuals in $\bX'$ and the original input $\bx$. This is computed as:
\begin{align*}
    \text{Proximity Score} = \frac{1}{B}\sum_{b=1}^B \text{distance} (\bx'_{b}, \bx)
\end{align*}
where $\text{distance}(\cdot, \cdot)$ is a distance function between two instances that measures the fraction of $N$ columns or values that do not match.
\item \textbf{Diversity Score.}
We compute the diversity score of the generated counterfactuals in $\bX'$ as the mean of the distances between each pair of samples.
\begin{align*}
    \text{Diversity Score} = \frac{2}{B(B-1)}\sum_{i=1}^{B-1} \sum_{j=i+1}^B \text{distance}(\bx'_{i}, \bx'_{j}) 
\end{align*}
where $\text{distance}(\cdot, \cdot)$ is a distance function between two instances that measures the fraction of $N$ columns or values that do not match.
\item \textbf{Plausibility.}
The goal is to evaluate how likely is the generated counterfactual under the true data distribution. We learn a model of the desired distribution by learning an auto-regressive model $p_\phi$ over the tokens or values in the instances. This auto-regressive model is described in further detail in the supplementary material. To compute the plausibility score, we compute the negative log-likelihood of each generated counterfactual $\bx'_b \in \bX'$ using $p_\phi$. 
\begin{align*}
    \text{Plausibility} &= - \frac{1}{B}\sum_{b=1}^B \log p_\phi(\bx'_{b}) \\ &= - \frac{1}{B}\sum_{b=1}^B \sum_{n=1}^N \log p_\phi(\bx'_{b,n} | \bx'_{b,1}, \ldots, \bx'_{b, n-1})
\end{align*}
where a lower negative log-likelihood is desired for a more plausible counterfactual.
\end{enumerate}
\begin{figure*}[t]
    \centering
    \includegraphics[width=\textwidth]{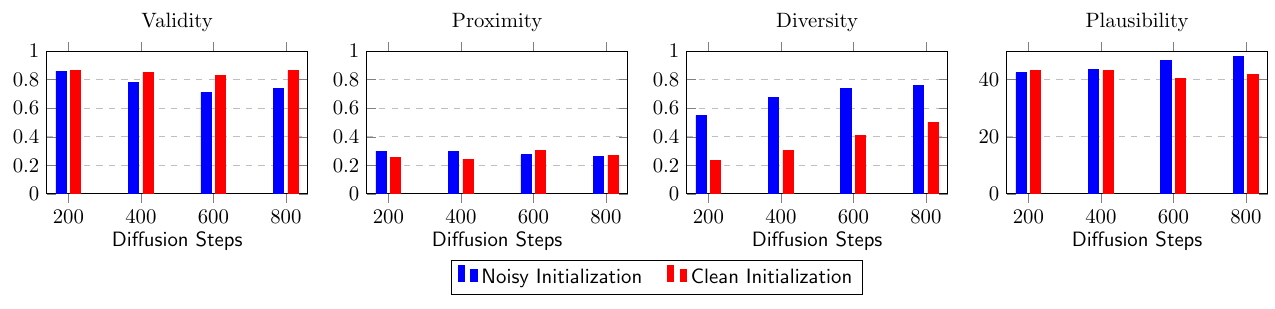}
    \caption{\textbf{Changing Diffusion Steps.} We report our metrics with respect to 1) changing diffusion steps, and 2) starting the guided diffusion with (shown in blue) and without (shown in red) adding an initial noise input. 
    We note that SCD remains robust to varying diffusion steps. Furthermore, we note a remarkable drop in diversity when the initial noise is not added at the start of the guided diffusion process.
     }
    \label{fig:effect-of-diff-steps}
\end{figure*}

\begin{figure*}
    \centering
    \includegraphics[width=0.48\textwidth]{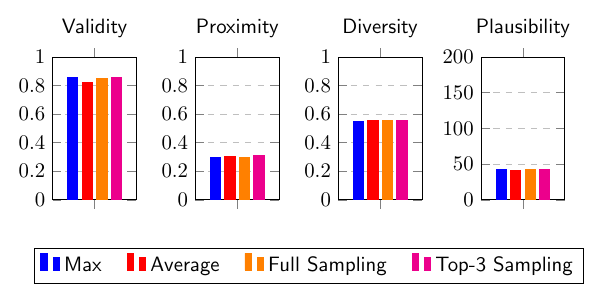} 
    \hfill\includegraphics[width=0.48\textwidth]{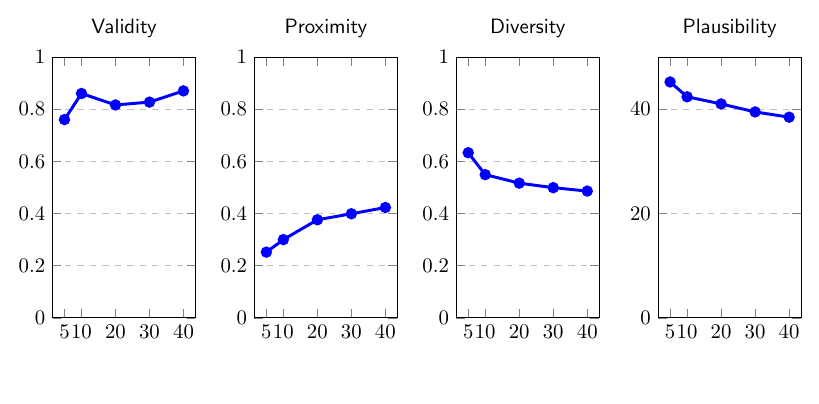}
    \caption{\textbf{Left: Effect of sampling strategies.} We vary the sampling strategies in our guided diffusion process and show the effect on our metrics. We observe a slightly higher validity score for \texttt{Max}. For other metrics, the scores remain robust. \textbf{Right: Effect of Choice of $B$.} We report our metric with respect to $B$, the number of counterfactuals generated.}
    \label{fig:sampling-strategy}
\end{figure*}

\begin{table*}[t]
    \scriptsize
    \centering
    \caption{\textbf{Counterfactual Samples in Adult Income Dataset.} Given the input row with the original label ``$\leq50K$", we ask our method SCD to generate counterfactual instances that flip the label to ``$>50K$" with respect to a black-box income predictor. We note that SCD generates plausible samples.}
    \label{tab:quals}
    \begin{tabular}{p{0.6cm} p{0.8cm} p{1.2cm} p{1.1cm} p{0.6cm} p{1.8cm} p{1.7cm} p{1.5cm} p{0.8cm} p{0.7cm} p{0.6cm} p{1.2cm}}
        \toprule
        \textbf{Method} & \textbf{Age} & \textbf{Workclass} & \textbf{Education} & \textbf{Ed. No.} & \textbf{Marital Status} & \textbf{Occupation} & \textbf{Relationship} & \textbf{Race} & \textbf{Gender} & \textbf{Hr/W} & \textbf{Country} \\
        \midrule
        \textbf{Input} & 38 & Private & HS-grad & 9 & Divorced & Handlers-cleaners & Not-in-family & White & Male & 40 & United-States \\
        \midrule
        \textbf{Ours} & 38 & Private & 9th & 5 & Never-married & Handlers-cleaners & Other-relative & White & Male & 40 & United-States \\
         & 38 & Private & HS-grad & 9 & Divorced & Other-service & Not-in-family & Black & Female & 40 & United-States \\
         & 44 & Private & HS-grad & 9 & Divorced & Handlers-cleaners & Unmarried & White & Female & 40 & United-States \\
        \midrule
        \textbf{Input} & 49 & Private & 9th & 5 & Married-spouse-absent & Other-service & Not-in-family & Black & Female & 16 & Jamaica \\
        \midrule
        \textbf{Ours} & 31 & Private & 5th-6th & 3 & Married-spouse-absent & Other-service & Not-in-family & Other & Female & 35 & Jamaica \\
         & 61 & Private & 9th & 5 & Never-married & Machine-op-inspct & Other-relative & Black & Female & 16 & Trinadad\&Tobago \\
         & 49 & Private & 9th & 5 & Separated & Other-service & Not-in-family & Black & Female & 48 & Jamaica \\
        \midrule
        \textbf{Input} & 23 & Private & Bachelors & 13 & Never-married & Adm-clerical & Own-child & White & Female & 30 & United-States \\
        \midrule
        \textbf{Ours} & 23 & Private & Bachelors & 13 & Never-married & Adm-clerical & Own-child & Black & Female & 30 & United-States \\
         & 25 & Private & Bachelors & 13 & Never-married & Adm-clerical & Own-child & White & Female & 30 & United-States \\
         & 23 & Private & Bachelors & 13 & Never-married & Adm-clerical & Own-child & White & Female & 30 & South \\
        \midrule
        \textbf{Input} & 39 & State-gov & Bachelors & 13 & Never-married & Adm-clerical & Not-in-family & White & Male & 40 & US \\
        \midrule
        \textbf{Ours} & 31 & Self-emp-inc & Bachelors & 13 & Married-civ-spouse & Adm-clerical & Wife & White & Female & 40 & US \\
         & 34 & Self-emp-inc & Bachelors & 13 & Married-civ-spouse & Exec-managerial & Husband & White & Male & 55 & US \\
         & 39 & Federal-gov & Bachelors & 13 & Married-civ-spouse & Prof-specialty & Husband & White & Male & 40 & US \\
        \midrule
        \textbf{Input} & 49 & Private & HS-grad & 9 & Married-spouse-absent & Craft-repair & Husband & White & Male & 40 & United-States \\
        \midrule
        \textbf{Ours} & 49 & Private & HS-grad & 9 & Married-spouse-absent & Craft-repair & Own-child & White & Male & 40 & Canada \\
         & 27 & Private & HS-grad & 9 & Married-civ-spouse & Other-service & Other-relative & Amer-Indian-Eskimo & Female & 48 & United-States \\
         & 49 & Private & HS-grad & 9 & Separated & Craft-repair & Not-in-family & White & Male & 55 & United-States \\
        \midrule
        \textbf{Input} & 19 & Private & HS-grad & 9 & Married-AF-spouse & Adm-clerical & Wife & White & Female & 25 & United-States \\
        \midrule
        \textbf{Ours} & 19 & Private & 9th & 5 & Never-married & Adm-clerical & Own-child & White & Female & 25 & United-States \\
         & 18 & Private & HS-grad & 9 & Never-married & Adm-clerical & Own-child & Black & Female & 20 & United-States \\
         & 18 & Private & HS-grad & 9 & Never-married & Adm-clerical & Own-child & White & Female & 25 & Canada \\
        \bottomrule
    \end{tabular}
\end{table*}

\subsection{Benefits of SCD in Counterfactual Generation}

In Table \ref{tab:quantitative-results-scd}, we compare our model SCD and our baseline, DiCE. It is remarkable that our model produces counterfactuals that are significantly more plausible than those generated by DiCE. In fact, the negative log-likelihood of our samples are 21.21, 42.37, and 42.91 while DiCE yields significantly worse results attaining 121.0, 166.7, and 109.5 on the 3 datasets, respectively. Our higher plausibility is also evidenced by our generated counterfactual samples in Table \ref{tab:qual}. We can see that our model coherent values for the columns \textit{Marital Status} and \textit{Relationship} while the baseline DiCE produces contradictory values e.g., \textit{Divorced} and \textit{Husband} within the same row.
This highlights the advantage of using a diffusion model that learns complex relationships to constrain the generated counterfactuals to be plausible. We show additional qualitative counterfactual samples generated by SCD in Table \ref{tab:quals}.

Furthermore, our results show significant improvements in the diversity and proximity scores over the baseline, achieving approximately 0.10-0.17 higher diversity and 0.04-0.10 higher proximity scores relative to DiCE. Our validity score, i.e., the fraction of generated counterfactuals that attain the desired label, is about $0.1$ lower than the baseline. While this is a slight decline, it is not a significant concern since it is straightforward to remove the counterfactuals that do not attain the desired label via post-processing. Furthermore, some worsening of the validity score may be expected since SCD constrains the samples to be plausible while DiCE does not. 
% Thus, ours creates a more difficult optimization problem to solve during the guided diffusion. 

\subsection{Analysis of Model Characteristics}
% In this section, we analyze questions regarding various characteristics of our model.\\

\noindent
\textbf{Question 1. How does dropping various loss terms affect performance?}\\

\noindent
In our guiding loss, we used 3 terms: validity, proximity, and diversity. While our default version retains all three terms, we would like to assess what would happen if each of the three terms were individually dropped. In Fig.~\ref{fig:ablation}, we report these results. 

\begin{enumerate}
    \item \textit{Dropping the Validity Term.} When we drop the validity term, we note that the validity score drops close to 0. That is, no generated samples are actually able to achieve the counterfactual label. This observation is shared for both our model as well as the baseline. This can be expected since the validity term in the loss is the only way to inform the generation process about the disparity between desired and the predicted label. Furthermore, all generated samples collapse to the original input, as suggested by a high proximity score and a very low diversity score.
    \item \textit{Dropping the Proximity Term.} When we drop the proximity term, we note that the scores are not significantly affected. We think this is because, in both SCD and the baseline, the process of generating the counterfactual starts with the original input, and the update steps are not able to deviate significantly from the original input.
    \item \textit{Dropping the Diversity Term.} When we drop the diversity term, we note, remarkably, that the diversity of samples of DiCE drops to 0. In comparison, our counterfactuals maintain high diversity even after removing the diversity loss term. This shows a unique characteristic of our model that by leveraging stochastic denoising, the samples of our model become naturally diverse. On the other hand, the existing models lack such stochasticity, requiring an explicit diversity loss term and careful tuning of its coefficient.
\end{enumerate}

\noindent
\textbf{Question 2. How does varying the number of guided diffusion steps affect performance?}\\

\noindent
We perform this analysis and report results in Fig.~\ref{fig:effect-of-diff-steps}. We note that our model is remarkably robust to the number of guided diffusion steps in terms of validity, proximity, and plausibility. In the diversity score, however, we see a slight upward trend with the increasing number of diffusion steps. We think this is because the diffusion steps are stochastic. Thus, accumulating randomness from a greater number of diffusion steps appears to promote higher sample diversity.\\

\noindent
\textbf{Question 3. How does adding noise at the start of guided diffusion affect performance?}\\

\noindent
In Fig.~\ref{fig:effect-of-diff-steps}, we also compare our model with and without adding noise at the start of the guided diffusion. We note that adding the noise is clearly beneficial since not adding the noise worsens the diversity score. This observation is consistent across different numbers of diffusion steps during counterfactual generation.\\

\noindent
\textbf{Question 4. How does varying sampling strategy for guided diffusion affect performance?}\\

\noindent
We test various sampling strategies during guided diffusion and whether it affects the performance or not. We test 4 sampling strategies for the denoising diffusion step. The first strategy is to choose the highest probability embedding per column (denoted as \texttt{Max}). The second strategy is to use a probability-weighted average of embeddings (denoted as \texttt{Average}). The third strategy is to sample an embedding under the predicted distribution (denote as \texttt{Full Sampling}). Lastly, our fourth strategy is to take the top-3 highest probability embeddings and randomly sample among these. Across all 4 strategies, in Fig.~\ref{fig:sampling-strategy}, we find the performances to be similar, indicating that our model is robust to this choice.\\

% We assess the influence of changing sampling strategies in Figure \ref{fig:sampling-strategy} and observe that "max" strategy exhibits a slightly higher validity score. 
\noindent
\textbf{Question 5. How does the number of generated counterfactuals affect performance?}\\

\noindent
We vary the number of generated counterfactuals in parallel (denoted as $B$) and report performance in Figure \ref{fig:sampling-strategy}. Note that we did not re-tune or change any hyperparameters other than $B$. We note that our performance remains robust with this change across all metrics. 
% Furthermore, we assessed the \textbf{influence of changing sampling strategies} through yet another experiment, presented in Figure \ref{fig:sampling-strategy}, using the UCI dataset. We observed a slightly higher validity for the "max" strategy, while other metrics remained largely stable.
% \begin{figure*}
%     \centering
%     \includegraphics[width=0.50\textwidth]{diagrams/changing-cf.pdf}
%     \caption{\textbf{Effect of Choice of $B$.} We report our metric with respect to $B$, the number of counterfactuals generated.
%     % We observe that there is an upward trend seen in proximity and plausibility as we increase the number of counterfactuals. In the case of diversity, we see a downward trend. For validity, it generally remains stable.
%     }
%     \label{fig:no-of-cf}
% \end{figure*}
% Lastly, we investigated the \textbf{effects of altering the number of generated counterfactuals}, as shown in Figure \ref{fig:no-of-cf} using the UCI dataset. Our findings indicate that increasing the number of counterfactuals results in higher proximity but decreased diversity. Plausibility improves with more counterfactuals, while the validity metric does not exhibit a clear trend in response to these changes.

\section{Conclusion}
In this paper, we introduced a novel counterfactual explainer called \textit{Structured Counterfactual Diffuser} (SCD) for structured data aimed at producing highly plausible counterfactuals. Our technique leverages a diffusion model to learn complex relationships among various attributes of structured data. Via guided diffusion, our model not only exhibits high plausibility compared to the existing state-of-the-art but also shows significant improvement in proximity and diversity, while also maintaining high validity.  In our analysis, we thoroughly analyze various important aspects of our proposed model, revealing useful insights. We find that our method removes the need for an explicit diversity loss by utilizing stochastic denoising that naturally produces diverse samples.
% substantiate the efficacy and robustness of our approach through various ablations, highlighting its benefits.

% \begin{IEEEkeywords}
% component, formatting, style, styling, insert
% \end{IEEEkeywords}

\bibliographystyle{IEEEtran}
\bibliography{satml}

\clearpage

\begin{appendices}
\begin{table*}[]
\centering
% \usepackage{booktabs}
% \usepackage{multirow}
\begin{tabular}{@{}|l|l|lll|@{}}
\toprule
\multicolumn{1}{|c|}{\multirow{2}{*}{\textbf{Model}}} & \multicolumn{1}{c|}{\multirow{2}{*}{\textbf{Hyperparameters}}}                                                                                                                                                               & \multicolumn{3}{c|}{\textbf{Dataset}}                                                                                                                                                                                                                                                                                                                 \\ \cmidrule(l){3-5} 
\multicolumn{1}{|c|}{}                       & \multicolumn{1}{c|}{}                                                                                                                                                                                               & \multicolumn{1}{l|}{\textbf{Adult Income}}                                                                                    & \multicolumn{1}{l|}{\textbf{UCI Bank}}                                                                                        & \textbf{House Price }                                                                                   \\ \midrule
Diffusion LM Pre-training                    & \begin{tabular}[c]{@{}l@{}}Batch Size\\ \# Epochs\\ Max Text Length\\ \# Diffusion Steps\\ Learning Rate\\ \# Learning Rate Warmup Steps\\ \# Learning Rate Half Life\\ Gradient Clipping\end{tabular} & \multicolumn{1}{l|}{\begin{tabular}[c]{@{}l@{}}120\\ 500\\ 11\\ 2000\\ 1e-4\\ 30000\\ 25000\\ 0.05\end{tabular}} & \multicolumn{1}{l|}{\begin{tabular}[c]{@{}l@{}}120\\ 500\\ 16\\ 2000\\ 1e-4\\ 30000\\ 25000\\ 0.05\end{tabular}} & \begin{tabular}[c]{@{}l@{}}120\\ 500\\ 9\\ 2000\\ 1e-4\\ 30000\\ 25000\\ 0.05\end{tabular} \\ \midrule
Guided Diffusion                            & \begin{tabular}[c]{@{}l@{}}\# Classes\\ Weight Coefficient of Proximity Loss\\ Weight Coefficient of Validity Loss\\ Weight Coefficient of Diversity Loss\\ Guider Learning Rate\\ Vocabulary Size\end{tabular}     & \multicolumn{1}{l|}{\begin{tabular}[c]{@{}l@{}}2\\ 0.01\\ 1.0\\ 0.001\\ 1.5\\ 2000\end{tabular}}                      & \multicolumn{1}{l|}{\begin{tabular}[c]{@{}l@{}}2\\ 0.01\\ 1.0\\ 0.001\\ 1.5\\ 5000\end{tabular}}                      & \begin{tabular}[c]{@{}l@{}}2\\ 0.01\\ 1.0\\ 0.001\\ 1.5\\ 5000\end{tabular}                     \\ \midrule
Plausibility Metric (GRU Model)                          & \begin{tabular}[c]{@{}l@{}}Batch Size\\ \# Epochs\\ Vocabulary Size\end{tabular}                                                                                                                       & \multicolumn{1}{l|}{\begin{tabular}[c]{@{}l@{}}120\\ 500\\ 2000\end{tabular}}                                    & \multicolumn{1}{l|}{\begin{tabular}[c]{@{}l@{}}120\\ 500\\ 5000\end{tabular}}                                    & \begin{tabular}[c]{@{}l@{}}120\\ 500\\ 5000\end{tabular}                                   \\ \midrule
DiCE                                         & \begin{tabular}[c]{@{}l@{}}\# Classes\\ Weight Coefficient of Proximity Loss\\ Weight Coefficient of Validity Loss\\ Weight Coefficient of Diversity Loss\\ Guider Learning Rate\\ Vocabulary Size\end{tabular}     & \multicolumn{1}{l|}{\begin{tabular}[c]{@{}l@{}}2\\ 0.1\\ 1.0\\ 0.0325\\ 2.5\\ 2000\end{tabular}}                     & \multicolumn{1}{l|}{\begin{tabular}[c]{@{}l@{}}2\\ 0.1\\ 1.0\\ 0.0325\\ 2.5\\ 5000\end{tabular}}                     & \begin{tabular}[c]{@{}l@{}}2\\ 0.1\\ 1.0\\ 0.0325\\ 2.5\\ 5000\end{tabular}                    \\ \midrule
Wachter                                      & \begin{tabular}[c]{@{}l@{}}\# Classes\\ Weight Coefficient of Proximity Loss\\ Weight Coefficient of Validity Loss\\ Guider Learning Rate\\ Vocabulary Size\end{tabular}                                            & \multicolumn{1}{l|}{\begin{tabular}[c]{@{}l@{}}2\\ 0.1\\ 1.0\\ 2.5\\ 2000\end{tabular}}                              & \multicolumn{1}{l|}{\begin{tabular}[c]{@{}l@{}}2\\ 0.1\\ 1.0\\ 2.5\\ 5000\end{tabular}}                              & \begin{tabular}[c]{@{}l@{}}2\\ 0.1\\ 1.0\\ 2.5\\ 5000\end{tabular}                             \\ \bottomrule
\end{tabular}
\vspace{0.6mm}
\caption{Hyperparameters of our model used in our experiments.}
    \label{tab:hyperparameters}
\end{table*}

\section{Implementation Details}
In this section, we provide additional implementation details. We also provide the exact hyperparameters we used in our experiments in Table \ref{tab:hyperparameters}.

\subsection{Diffusion Model Pre-Training}
\label{diffusion-pretraining}

\noindent
\textbf{Building Row Embeddings.} We transform a table row into a vector. For numeric columns or values, we first discretize them via binning into equal sized bins. Then, for the discretized numerical columns as well as the categorical columns, their discrete value is represented as an integer. For each column, we maintain a learned embedding dictionary. Corresponding to the integer value of each column, we retrieve the corresponding embedding from the learned embedding dictionary. The embeddings of all the columns are concatenated to create a vector representation of the entire row. The embeddings in the dictionary are learned during the training of the diffusion model and frozen afterwards.

\subsection{Details of Guided Diffusion for Counterfactual Generation}

\noindent
\textbf{Loss Coefficients.} For the guiding objective function, $\lambda_\text{validity}$ is set to 1.0, $\lambda_\text{proximity}$ is set to 0.01 and $\lambda_\text{diversity}$ is set to 0.0001. Note that our method works with the same coefficients for all datasets, providing evidence of our method's robustness.

\noindent
\textbf{Classifier.} The classifier is implemented as a simple 2-layer MLP with hidden dimension 768. It takes the row embeddings as input and predicts the class label depending on the task associated with each dataset. For each dataset, their corresponding MLP classifier are trained using a cross-entropy loss.

% \subsection{Inference}
% The hyperparameters are listed in Table \ref{tab:hyperparameters}.
% \textbf{Hyperparameters.} The seed hyperparameter is set to 0, ensuring reproducibility of results by initializing random number generators with the same seed value across different runs. The batch\_size is defined as 10, determining the number of data samples included in each iteration of training.

% The max\_text\_len hyperparameter is configured as the number of features in a dataset, representing the maximum allowable length for input text sequences. The diff\_steps hyperparameter is set to 2000, indicating the number of steps between consecutive differences used in the diffusion process. The vocab\_size is set to 5000, which determines the size of the vocabulary used in text-related tasks.

% The guider\_label hyperparameter is set to 0, representing the class label for the guider component. The guider\_lr hyperparameter is set to 40.0, denoting the learning rate used for the guider's optimization process. The guider\_path specifies the path to the pre-trained model checkpoint for the guider. The plausibility\_path points to the path of the pre-trained model checkpoint for plausibility assessment. The guider\_num\_classes is set to 2, indicating the number of classes in the guider component's classification task.

% The num\_rounds hyperparameter is set to 1, indicating the number of rounds in the experiment. The mode is configured as 'hard\_max', possibly determining the sampling mode of the diffusion process.

\subsection{Plausibility Metric}
In this section, we describe how we compute the plausibility metric. To measure plausibility, we compute the negative log-likelihood of the generated counterfactuals. The log-likelihood is computed with respect to the estimated data distribution modeled via an autoregressive RNN and another autoregressive Transformer model. Note that these architectures share no inductive biases or parameters with the models we evaluate, and thus, can be considered as objective measures.

\noindent
\textbf{RNN Model.} The RNN model is trained on the tabular data by asking to recurrently predict the values within each row sequentially from left to right. The RNN is trained via teacher-forcing and cross-entropy loss for each value. The hidden dimension of the RNN model is 768.

\noindent
\textbf{Transformer Model.} The transformer model is trained on the tabular data by asking it to predict (under causal masking) the values within each row. This essentially makes it an autoregressive model of the row. The transformer is trained to predict each row value via a cross-entropy loss conditioned on the values on the left. The hidden dimensions of the transformer model is 768. It is a transformer with 4 layers and 4 heads.

\subsection{Additional Results}

Here, we provide some additional experiment results and analyses.

\textbf{Evaluation of Valid-Only Counterfactuals.} In this section, we computed our metrics i.e., proximity, diversity and plausibility with only valid counterfactuals. We show these results in Table \ref{tab:valid-results-scd}. We find that the performance trend is similar to the trend noted without filtering away the non-valid counterfactuals, with our model SCD outperforming the baselines.

\begin{table*}[t]
% \scriptsize
\centering
  \caption{Comparison of plausibility, proximity, diversity scores of SCD, DiCE and Wachter on various datasets with only valid counterfactuals. For proximity and diversity scores, higher is better. For the plausibility score, lower is better since it captures the negative log-likelihood of the generated samples.}
    \label{tab:valid-results-scd}
\begin{tabular}{lccccccc}
\toprule
\multicolumn{1}{l}{\textbf{Dataset}} & \multicolumn{3}{c}{\textbf{Plausibility} $(\downarrow)$} & \multicolumn{3}{c}{\textbf{Proximity} $(\uparrow)$} \\
\cmidrule(lr){2-4} \cmidrule(lr){5-7}
 & \textbf{Wachter \textit{et al.}} & \textbf{DiCE} & \textbf{SCD} & \textbf{Wachter \textit{et al.}} & \textbf{DiCE} & \textbf{SCD} \\
\midrule
\textbf{Adult Income} & 110.57 & 120.10 & \textbf{21.99} & \textbf{0.677} & 0.581 & 0.583 \\
\textbf{UCI Bank} & 168.57 & 168.99 & \textbf{74.64} & 0.223 & 0.210 & \textbf{0.323} \\
\textbf{Housing Price} & 104.88 & 109.63 & \textbf{74.57} & \textbf{0.365} & 0.300 & 0.314 \\
\midrule
\multicolumn{1}{l}{\textbf{Dataset}} & \multicolumn{3}{c}{\textbf{Diversity} $(\uparrow)$} \\
\cmidrule(lr){2-4}
 & \textbf{Wachter \textit{et al.}} & \textbf{DiCE} & \textbf{SCD} \\
\cmidrule(lr){0-3}
\textbf{Adult Income} & 0.00 & 0.396 & \textbf{0.414}\\
\textbf{UCI Bank} & 0.187 & 0.538 & \textbf{0.549} \\
\textbf{Housing Price} & 0.0 & \textbf{0.639} & 0.534 \\
\bottomrule
\end{tabular}
\end{table*}

\begin{table*}[t]
% \scriptsize
\centering
  \caption{Comparison of plausibility, proximity, diversity, and validity scores of SCD, DiCE-VAE, DiCE and Wachter on various datasets. For validity, proximity, and diversity scores, higher is better. For the plausibility score, lower is better since it captures the negative log-likelihood of the generated samples.}
    \label{tab:quantitative-results-additional-baseline}
\begin{tabular}{lccccccccc}
\toprule
\multicolumn{1}{l}{\textbf{Dataset}} & \multicolumn{4}{c}{\textbf{Plausibility} $(\downarrow)$} & \multicolumn{4}{c}{\textbf{Proximity} $(\uparrow)$} \\
\cmidrule(lr){2-5} \cmidrule(lr){6-9}
 & \textbf{Wachter \textit{et al.}} & \textbf{DiCE} & \textbf{DiCE-VAE}  & \textbf{SCD} & \textbf{Wachter \textit{et al.}} & \textbf{DiCE} & \textbf{DiCE-VAE}  &\textbf{SCD} \\
\midrule
\textbf{Adult Income} & 108.7 & 121.0 & 54.34 & \textbf{21.21} & 0.685 & 0.5764 & 0.623 & \textbf{0.6173} \\
% \textbf{UCI Bank} & 168.3 & 166.7 & TBA & \textbf{42.37} & 0.226 & 0.2141 & TBA & \textbf{0.3000} \\
\textbf{Housing Price} & 102.8 & 109.5 & 73.71 & \textbf{42.91} & 0.375 & 0.3055 & 0.337 & \textbf{0.3417} \\
\midrule
\multicolumn{1}{l}{\textbf{Dataset}} & \multicolumn{4}{c}{\textbf{Diversity} $(\uparrow)$} & \multicolumn{4}{c}{\textbf{Validity} $(\uparrow)$} \\
\cmidrule(lr){2-5} \cmidrule(lr){6-9}
 & \textbf{Wachter \textit{et al.}} & \textbf{DiCE} & \textbf{DiCE-VAE}  & \textbf{SCD} & \textbf{Wachter \textit{et al.}} & \textbf{DiCE} & \textbf{DiCE-VAE}  & \textbf{SCD} \\
\midrule
\textbf{Adult Income} & 0.002 & 0.3837 & 0.305 &\textbf{0.4008} & 0.9400 & \textbf{0.9776} & 0.847 & 0.7511 \\
% \textbf{UCI Bank} & 0.041 & 0.4165 & TBA &\textbf{0.5498} & 0.9900 & \textbf{0.9686} & TBA & 0.8600 \\
\textbf{Housing Price} & 0.03 & 0.4289 & 0.607 & \textbf{0.5986} & 0.9999 & \textbf{0.9908} & 0.855 & 0.8526 \\
\bottomrule
\end{tabular}
\end{table*}

\begin{table*}[t]
\centering
  \caption{Comparison of plausibility scores of SCD, DiCE-VAE, DiCE and Wachter on various datasets with GRU model and a transformer model. For the plausibility score, lower is better since it captures the negative log-likelihood of the generated samples.}
    \label{tab:quantitative-results-plausibility}
\begin{tabular}{lcccccccc}
\toprule
\multicolumn{1}{l}{\textbf{Dataset}} & \multicolumn{3}{c}{\textbf{Plausibility with GRU Model} $(\downarrow)$} & \multicolumn{4}{c}{\textbf{Plausibility with Transformer Model} $(\downarrow)$} \\
\cmidrule(lr){2-4} \cmidrule(lr){5-8}
 & \textbf{Wachter \textit{et al.}} & \textbf{DiCE} & \textbf{SCD} & \textbf{Wachter \textit{et al.}} & \textbf{DiCE} & \textbf{SCD} \\
\midrule
\textbf{Adult Income} & 108.7 & 121.0 & \textbf{21.21} & 85.18 & 94.70 & \textbf{28.07} \\
\textbf{UCI Bank} & 168.3 & 166.7 & \textbf{42.37} & 191.06 & 190.81 & \textbf{107.69} \\
\textbf{Housing Price} & 102.8 & 109.5 & \textbf{42.91} & 91.83 & 92.23 & \textbf{49.88} \\
\bottomrule
\end{tabular}
\end{table*}

\textbf{Incorporating Plausibility without Diffusion.}
In this section, we ask whether other traditional distribution modeling approaches e.g., VAEs, can also provide benefits in improving plausibility of the current state of the art or not? If yes, how does it compare with the use of diffusion modeling.

To test this, we created a model that we call DiCE-VAE. In DiCE-VAE, we train a VAE model to capture the data distribution in the row-embedding space. In the gradient search objective (similar to that of DICE), we simply add another term: negative ELBO or the Evidence Lower-Bound as estimated by the trained VAE model. We hypothesize that this additional loss term will prevent the search from exiting the plausible regions of the search space. The new guiding loss $\cL$ can be formally described as:
\begin{align*}
    \cL(\bZ', \bx, f, y') &= \lambda_\text{validity}\cL_\text{validity}(\bZ', f, y')\\ &+ \lambda_\text{proximity}\cL_\text{proximity}(\bZ, \bZ')\\ &+ \lambda_\text{diversity}\cL_\text{diversity}(\bZ')\\ &+ \lambda_\text{plausibility}\cL_\text{plausibility}(\bZ')
\end{align*}

In experiments, we find that this indeed improves the plausibility in comparison to the baseline DICE, almost halving the negative log-likelihood of the generated counterfactuals from DiCE-VAE. We show the results in Table \ref{tab:quantitative-results-additional-baseline}. 
We note that plausibility improves to 54.34 in DiCE-VAE as compared to Wachter and DiCE where it is  108.7 and 121.0 in Adult Income dataset. A similar trend is seen across three datasets as shown in Table \ref{tab:quantitative-results-additional-baseline}. 
However, comparing with SCD using diffusion model performs even better than DiCE-VAE, thus justifying the use of diffusion model over the traditional distribution modeling approaches e.g., VAEs. 

% We show another baseline DiCE-VAE which incorporates plausibility into the gradient search objective. The hypothesis is to investigate that adding such a plausibility term in the loss function in a gradient objective is not sufficient to introduce plausibility into counterfactual generation. 
% For implementation, we extend the guiding loss by adding an additional elbo loss computed by training a Variational Auto-encoder. 

% \subsection{Investigating different plausibility models.}
% To investigate our model \texttt{SCD's} performance on plausibility, we vary the RNN model to a transformer model and then compare and contrast the negative log-likelihood score from both of these models. We report these results in Table \ref{tab:quantitative-results-plausibility}. 
% We note that ...

% \usepackage{booktabs}
% \usepackage{multirow}
% \usepackage{longtable}
% Note: It may be necessary to compile the document several times to get a multi-page table to line up properly
% \section{Qualitative Examples}

\subsection{Ethics Statement}
Building language models with steering ability can help in reducing bias, toxicity, etc. Our proposed system SCD does not support or amplify any biases and can not be exploited to generate such content. Infact, it helps in generating counterfactuals that indeed aid in making the models more explainable and bias-free. Hence, this work poses no threat of discrimination, or bias. 
\end{appendices}

\end{document}